\documentclass{article}

\usepackage[numbers]{natbib}
\usepackage{subcaption}
\usepackage{enumitem}
\usepackage{booktabs}
\usepackage{amssymb}
\usepackage{subcaption}
\usepackage{booktabs}
\usepackage{times}
\usepackage{url}
\usepackage{hyperref}
\usepackage{natbib}
\usepackage{lscape}
\usepackage{amsmath}
\usepackage{graphicx}
\usepackage{arxiv}
\usepackage{xcolor}
\usepackage{algorithm}
\usepackage{algpseudocode}
\usepackage{amssymb}
\usepackage{listings}

\lstdefinestyle{pytorchstyle}{
  basicstyle=\ttfamily\footnotesize,
  backgroundcolor=\color{white},
  commentstyle=\color{gray},
  keywordstyle=\color{black},
  frame=single,
  columns=fullflexible,
  keepspaces=true,
  breaklines=true,
  breakatwhitespace=true,
  showstringspaces=false,
  language=Python,
  morecomment=[l]{\#}
}

\usepackage{tabularx}
\usepackage[justification=centering]{caption}
\graphicspath{{Image/}}

\def\tsc#1{\csdef{#1}{\textsc{\lowercase{#1}}\xspace}}
\tsc{WGM}
\tsc{QE}

\begin{document} 

\title{G-MSGINet: A Grouped Multi-Scale Graph-Involution Network for Contactless Fingerprint Recognition}

\author{
 Santhoshkumar Peddi \\
  Department of Computer Science and Engineering\\
  Indian Institute of Technology\\
  Kharagpur, India 721302 \\
  \texttt{santhoshpps11@gmail.com} \\
   \And
 Soham Bandyopadhyay \\
  Advanced Technology Development Centre\\
  Indian Institute of Technology\\
  Kharagpur, India 721302 \\
  \texttt{sohamban@gmail.com} \\
  \And
 Monalisa Sarma \\
  Subir Chowdhury School of Quality and Reliability\\
  Indian Institute of Technology\\
  Kharagpur, India 721302 \\
  \texttt{monalisa@iitkgp.ac.in }
  \And
 Debasis Samanta \\
  Department of Computer Science and Engineering\\
  Indian Institute of Technology\\
  Kharagpur, India 721302 \\
  \texttt{dsamanta@iitkgp.ac.in}
}

\maketitle

\begin{abstract}
This paper presents G-MSGINet, a unified and efficient framework for robust contactless fingerprint recognition that jointly performs minutiae localization and identity embedding directly from raw input images. Existing approaches rely on multi-branch architectures, orientation labels, or complex preprocessing steps, which limit scalability and generalization across real-world acquisition scenarios. In contrast, the proposed architecture introduces the G-MSGI layer, a novel computational module that integrates grouped pixel-level involution, dynamic multi-scale kernel generation, and graph-based relational modelling into a single processing unit. Stacked G-MSGI layers progressively refine both local minutiae-sensitive features and global topological representations through end-to-end optimization. The architecture eliminates explicit orientation supervision and adapts graph connectivity directly from learned kernel descriptors, thereby capturing meaningful structural relationships among fingerprint regions without fixed heuristics. Extensive experiments on three benchmark datasets, namely PolyU, CFPose, and Benchmark 2D/3D, demonstrate that G-MSGINet consistently achieves minutiae F1-scores in the range of $0.83 \pm 0.02$ and Rank-1 identification accuracies between $97.0\%$ and $99.1\%$, while maintaining an Equal Error Rate (EER) as low as $0.5\%$. These results correspond to improvements of up to $4.8\%$ in F1-score and $1.4\%$ in Rank-1 accuracy when compared to prior methods, using only $0.38$ million parameters and $6.63$ giga floating-point operations, which represents up to ten times fewer parameters than competitive baselines. This highlights the scalability and effectiveness of G-MSGINet in real-world contactless biometric recognition scenarios.

\textbf{Keywords} - Contactless Fingerprint Recognition, Minutiae detection, Graph neural networks, Multi-scale feature extraction, End-to-end fingerprint analysis
\end{abstract}

\section{Introduction}\label{intro}
Fingerprint recognition remains one of the most reliable biometric modalities due to its uniqueness, permanence, and user-friendly acquisition process \cite{jain1997line}. Traditional systems have predominantly relied on contact-based sensors that offer high accuracy but introduce significant practical limitations, including hygiene concerns, skin deformation, and inconsistent finger placement \cite{Survey}. These limitations became particularly evident during the COVID-19 pandemic, accelerating interest in contactless fingerprint recognition technologies that provide hygienic and touch-free authentication alternatives. Despite their advantages, contactless acquisition systems introduce new challenges: perspective distortion, variable lighting conditions, and diminished ridge-valley contrast, which significantly complicate the extraction of robust and discriminative features, particularly minutiae points, from raw images \cite{tan2020towards}.

Minutiae points, primarily ridge endings and bifurcations, are widely regarded as cornerstone features in fingerprint recognition due to their high discriminative power. Consequently, early efforts in contactless fingerprint recognition naturally extended minutiae-based methodologies, employing patch-based or heatmap-based techniques for their localization and orientation estimation \cite{tan2020towards,cotrim2022multiscale,cotrim2023residual,zhang2021multi}. These methods often rely on complex multi-stage or multi-branch architectures, which increase computational overhead and system complexity. However, a critical bottleneck limiting the effectiveness of these approaches lies in the lack of publicly available ground-truth minutiae annotations for contactless datasets. This scarcity of labeled data forces researchers to either engage in labor-intensive manual annotation or depend on off-the-shelf tools originally designed for contact-based fingerprints, such as VeriFinger \cite{neurotechnology2025} or MINDTCT \cite{ko2007user}. These tools frequently yield suboptimal results on contactless images unless accompanied by extensive and error-prone preprocessing \cite{vyas2024collaborative,birajadar2016touch}.

To circumvent these annotation challenges, texture-based approaches emerged that extract global or local patterns directly from fingerprint images without explicitly detecting minutiae points \cite{shi2022novel}. While more straightforward, these methods struggle with variable image quality, inconsistent lighting, and contrast issues, factors commonly encountered in uncontrolled environments. This limitation led to the development of minutiae-aware texture approaches that incorporate minutiae location and orientation information into the feature extraction process \cite{birajadar2016touch,yin2019contactless,attrish2021contactless,parasnis2024verifnet}. Though effective, these hybrid methods also implement multi-branch architectures with separate network paths for minutiae detection and feature representation, resulting in increased computational demands and architectural complexity.

In parallel, another research direction has emphasized cross-matching strategies that bridge the gap between contact-based and contactless modalities through domain adaptation, deformation modeling, or synthetic data generation. These approaches prioritize transforming contactless inputs to resemble contact-based features rather than developing representations specifically tailored to the unique characteristics of contactless fingerprints \cite{lin2018cnn,lin2018matching,grosz2021c2cl,9417198,hasan2022deep,dong2023synthesis}. This transformation-focused paradigm, while valuable for compatibility with existing databases, fails to fully leverage the distinct information content available in native contactless acquisitions.

The limitations in these research trajectories reveal several interconnected challenges. The reliance on minutiae orientation, particularly problematic given the difficulty of reliable orientation annotation in variable-quality contactless images, constrains the development of robust recognition systems. Current methods also exhibit limited modeling of spatial dependencies between minutiae points, neglecting the vital topological structure that carries discriminative information about fingerprint uniqueness \cite{neumann2015quantifying}. Furthermore, the absence of unified, orientation-free architectures specifically designed for raw contactless inputs forces reliance on complex preprocessing pipelines that introduce additional error sources \cite{priesnitz2021overview}. These interconnected limitations highlight a significant research gap: the absence of a structurally aware, unified framework capable of simultaneously performing minutiae localization and global representation learning directly from raw contactless images without requiring orientation labels, fixed graph connections, or modality transformation.

To address these challenges, the proposed approach sets out the following objectives:
\begin{itemize}
\setlength{\itemsep}{0pt}
	\setlength{\parskip}{0pt} 
	\setlength{\parsep}{0pt}
    \item Develop a unified single-branch architecture for simultaneous minutiae detection and identity embedding.
    \item Eliminate explicit minutiae orientation dependency and complex preprocessing requirements.
    \item Design adaptive local feature extraction on varying spatial scales to effectively capture fingerprint structures at the minutiae level and the ridge level.
    \item Integrate graph-based relational reasoning through learnable, data-driven mechanisms.
\end{itemize}

The proposed approach is guided by three key innovations, each addressing specific limitations of existing methods. First, adaptive kernel generation through involution \cite{li2021involution} operations forms the foundation of the architecture. Unlike standard convolutions that apply spatially invariant filters across the entire image, involution operations dynamically generate position-specific kernels conditioned on local content. This spatial adaptivity is crucial for fingerprint recognition as it enables the network to adjust its processing based on local ridge patterns, minutiae characteristics, and image quality variations. Through this content-adaptive processing, involution operations capture the intricate structural details that define fingerprint uniqueness with significantly higher precision than standard convolutions, while maintaining computational efficiency by sharing kernel-generating parameters across spatial locations.

Second, the multi-scale adaptive feature extraction mechanism captures the diverse spatial patterns present in fingerprint images. These patterns include both fine minutiae details and larger ridge structures, which are traditionally processed through separate paths with different receptive fields. Instead, the proposed kernel generation strategy produces scale-specific features within a single, unified framework. This design efficiently combines information across spatial levels, enhancing the accuracy of minutiae localization and the reliability of identity embeddings under varying image conditions.

Third, the data-driven graph construction process transforms how structural relationships between fingerprint regions are modeled. Traditional graph-based approaches rely on explicit minutiae extraction followed by fixed connection rules, such as k-nearest neighbors or Delaunay triangulation \cite{shi2022towards,yin2019contactless}, which often lack flexibility in adapting to the inherent variability of contactless inputs. In contrast, the proposed end-to-end graph learning mechanism infers connectivity directly from learned feature representations, allowing the model to uncover semantically meaningful relationships that extend beyond simple spatial proximity. By removing the dependence on predefined topological rules, this approach enables more adaptive and structurally aware modeling of fingerprint data, ultimately improving recognition robustness across diverse acquisition conditions.

Building on these innovations, this paper presents G-MSGINet, a Grouped Multi-Scale Graph-Involution Network specifically designed for robust contactless fingerprint recognition. The core architectural component is the G-MSGI layer, which seamlessly integrates grouped involution, multi-scale processing, and graph-based relational modeling into a unified computational module. By stacking multiple G-MSGI layers and jointly optimizing their parameters, the complete architecture simultaneously learns to detect minutiae points and extract discriminative fingerprint embeddings directly from raw images.

The primary contributions of this work include:
\begin{itemize}
\setlength{\itemsep}{0pt}
	\setlength{\parskip}{0pt} 
	\setlength{\parsep}{0pt}
    \item A novel G-MSGI layer that fuses spatially adaptive kernel generation with relational reasoning.
    \item A data-driven mechanism for constructing adaptive graphs using involution kernel projections.
    \item An efficient, end-to-end unified architecture for joint minutiae localization and fingerprint embedding.
    \item State-of-the-art performance across three contactless fingerprint datasets with reduced annotation requirements.
\end{itemize}

In summary, this paper presents a unified framework for contactless fingerprint recognition that addresses the fundamental limitations of existing approaches through adaptively learned feature extraction and graph-based structural modeling. The remainder of this paper is organized as follows: Section 2 reviews related literature; Section 3 details the proposed G-MSGINet architecture; Section 4 presents comprehensive experimental evaluations; and Section 5 concludes with a discussion of limitations and future research directions.

\section{Related Work} \label{related-work}
Prior works in this domain can broadly be categorized into three key directions: minutiae-based approaches, deep feature-based approaches, and cross-matching methods for interoperability between contactless and contact-based fingerprints.

\textit{Minutiae-Based Approaches:} Early efforts in minutiae extraction for contactless fingerprint recognition date back to Birajadar et al. \cite{birajadar2016touch}, who proposed a monogenic wavelet-based fingerphoto enhancement technique using phase congruency, significantly improving minutiae extraction under unconstrained image conditions. Yin et al. \cite{yin2019contactless} introduced a robust framework leveraging global minutiae topology along with a loose genetic algorithm, optimizing minutiae correspondence through an innovative similarity matrix that captured minutiae pairs and ridge count information. Tan et al. \cite{tan2020towards} subsequently developed a deep learning-based minutiae extraction framework followed by a three-stage pose compensation process involving core-point-based view estimation, ellipsoid modeling, and alignment procedures. Zhang et al. \cite{zhang2021multi} moved toward jointly learning minutiae location and orientation maps through a multi-task convolutional neural network with attention mechanisms, thus providing end-to-end minutiae detection without explicit preprocessing. Further improvements were made by Cotrim et al. in two related works: initially through a multiscale CNN approach designed specifically for low-contrast contactless images \cite{cotrim2022multiscale}, and later a residual squeeze-and-excitation U-Net architecture to enhance minutiae detection accuracy under challenging imaging conditions \cite{cotrim2023residual}. Most recently, Vyas et al. \cite{vyas2024collaborative} introduced the extraction and analysis of complementary ridge and valley minutiae features, along with exploring color-space transformations, achieving improvements over previous methods.

\textit{Deep Feature-Based Approaches:} Parallel research directions have investigated deep feature based methods that capture holistic fingerprint representations, either bypassing traditional minutiae extraction steps or incorporating them into the learning process. Early efforts such as Malhotra et al. \cite{malhotra2020matching} explored deep scattering network features in conjunction with random forest classification to enable finger-selfie authentication, demonstrating viability in both within-domain and cross-domain settings. Following this, Attrish et al. \cite{attrish2021contactless} introduced a hybrid Siamese CNN architecture that extracted global texture-based features and integrated local minutiae-based scores, balancing speed and accuracy for real-time authentication scenarios. The field then advanced toward more structured learning of minutiae features. Shi et al. \cite{shi2022towards} proposed a deep geometric graph neural network framework that combined CNN-based local feature extraction with graph-based relational modeling among minutiae points. Their architecture learned both low-level and geometric features, enabling binary classification of minutiae pairs and improving robustness to pose and lighting variations, as demonstrated across public datasets. Further extending this direction, Shi et al. \cite{shi2022novel} developed Fingerprint Triplet-GAN (FTG), an end-to-end architecture that combines triplet loss with adversarial learning to generate pose-invariant fingerprint embeddings without relying on explicit minutiae detection. Most recently, Parasnis et al. \cite{parasnis2024verifnet} presented VerifNet, which integrates scattering wavelet transforms, Siamese networks, and minutiae-level representations in a compact design to strike a balance between computational efficiency and accuracy.

\textit{Cross-Matching Approaches:} Considering the widespread presence of legacy contact-based fingerprint databases, cross-matching approaches aim to bridge the gap between contactless and contact-based modalities. Early contributions by Lin et al. include a multi-Siamese CNN-based system generating fingerprint representations using minutiae and ridge information combined with distance-aware loss functions \cite{lin2018cnn}, and a thin-plate spline (RTPS) deformation correction model to robustly align fingerprints across modalities \cite{lin2018matching}. Grosz et al. \cite{grosz2021c2cl} introduced C2CL, a comprehensive mobile-based framework combining preprocessing and matching strategies to effectively handle image variations caused by sensor quality and finger presentation. Subsequently, Tan et al. \cite{9417198} enhanced interoperability by proposing a dual-branch minutiae attention network trained using a reciprocal distance loss and complemented by pose compensation procedures. Hasan et al. \cite{hasan2022deep} employed a coupled conditional GAN-based deep learning framework, mapping contactless and contact-based images into a shared embedding space, thus reducing the domain gap. Priesnitz et al. \cite{priesnitz2022mobile} developed an automated smartphone-based contactless fingerprint recognition system, highlighting its practicality and comparable accuracy against traditional contact-based methods. Dong et al. \cite{dong2023synthesis} took a novel direction, synthesizing large-scale, multi-view 3D fingerprint databases for rigorous cross-modality evaluation, enhancing both accuracy and privacy aspects. Most recently, Artan et al. \cite{artan2024minnet} proposed a multi-scale approach explicitly addressing resolution and deformation variances between contact and contactless fingerprints.

While significant progress has been made in contactless fingerprint recognition through various minutiae-based, deep-feature, and cross-matching approaches, several crucial limitations persist. Most existing works involve extensive preprocessing steps, including explicit image enhancement, pose compensation, and normalization, which can introduce cumulative errors and reduce robustness, especially given the inherent low-contrast and variable pose of contactless fingerprints. Additionally, many state-of-the-art approaches rely on multi-stage or multi-branch pipelines, making these systems overly complex and difficult to optimize or interpret effectively. Another common limitation is their heavy reliance on minutiae orientation information, which is inherently noisy and less reliable in unconstrained contactless acquisitions. Furthermore, current literature predominantly focuses on cross-modality conversions, wherein contactless fingerprints are transformed or mapped to mimic the contact-based modality, rather than designing dedicated recognition solutions explicitly tailored to the unique characteristics of raw contactless fingerprint data. To the best of our knowledge, no existing approach provides a unified, structurally aware, and orientation-free framework specifically developed for raw contactless fingerprint inputs, highlighting a clear need for further exploration in this direction. Motivated by these gaps, we propose a novel unified graph-based representation learning paradigm that directly leverages raw contactless fingerprint images, eliminates dependence on minutiae orientation, and integrates minutiae detection and graph structural modeling within a streamlined, single-branch architecture. The details of this proposed approach are elaborated comprehensively in the next section.

\section{Proposed Method}
\begin{figure}[!ht]
    \centering
    \includegraphics[width=0.99\linewidth]{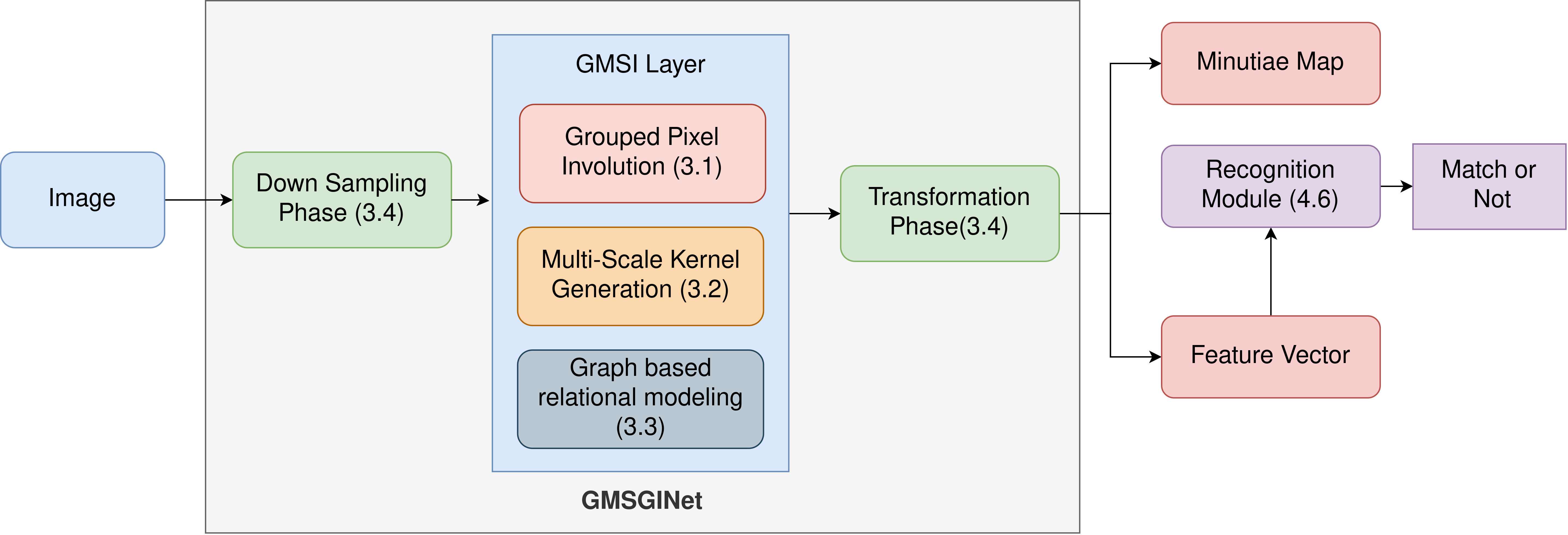}
    \caption{Overview of the proposed method. The network processes a full fingerprint image to generate a minutiae heatmap and an identity embedding. Numbered blocks correspond to key components, with section numbers indicating where each is described.}
    \label{fig:Overview}
\end{figure}

This section presents the design of the proposed G-MSGINet, a unified end-to-end architecture that processes a full contactless fingerprint image to jointly perform two tasks: (i) localization of minutiae points, and (ii) extraction of a compact identity embedding. The network generates two outputs from a single input image, a high-resolution minutiae likelihood map indicating potential minutiae locations and a minutiae-aware feature vector that captures global identity information. These representations support fingerprint recognition by enabling cosine similarity comparison between identity embeddings of different images. The core of G-MSGINet consists of stacked G-MSGI layers, each integrating three tightly coupled modules: grouped pixel-level involution for spatially adaptive local feature extraction, multi-scale dynamic kernel generation for capturing fine and coarse ridge structures, and graph-based relational modeling for learning structural relationships between spatial regions. These modules work collaboratively to enrich both local and global fingerprint representations. An overview of the full processing pipeline is illustrated in Fig.~\ref{fig:Overview}.

\begin{figure*}
    \centering
    \includegraphics[width=0.89\linewidth]{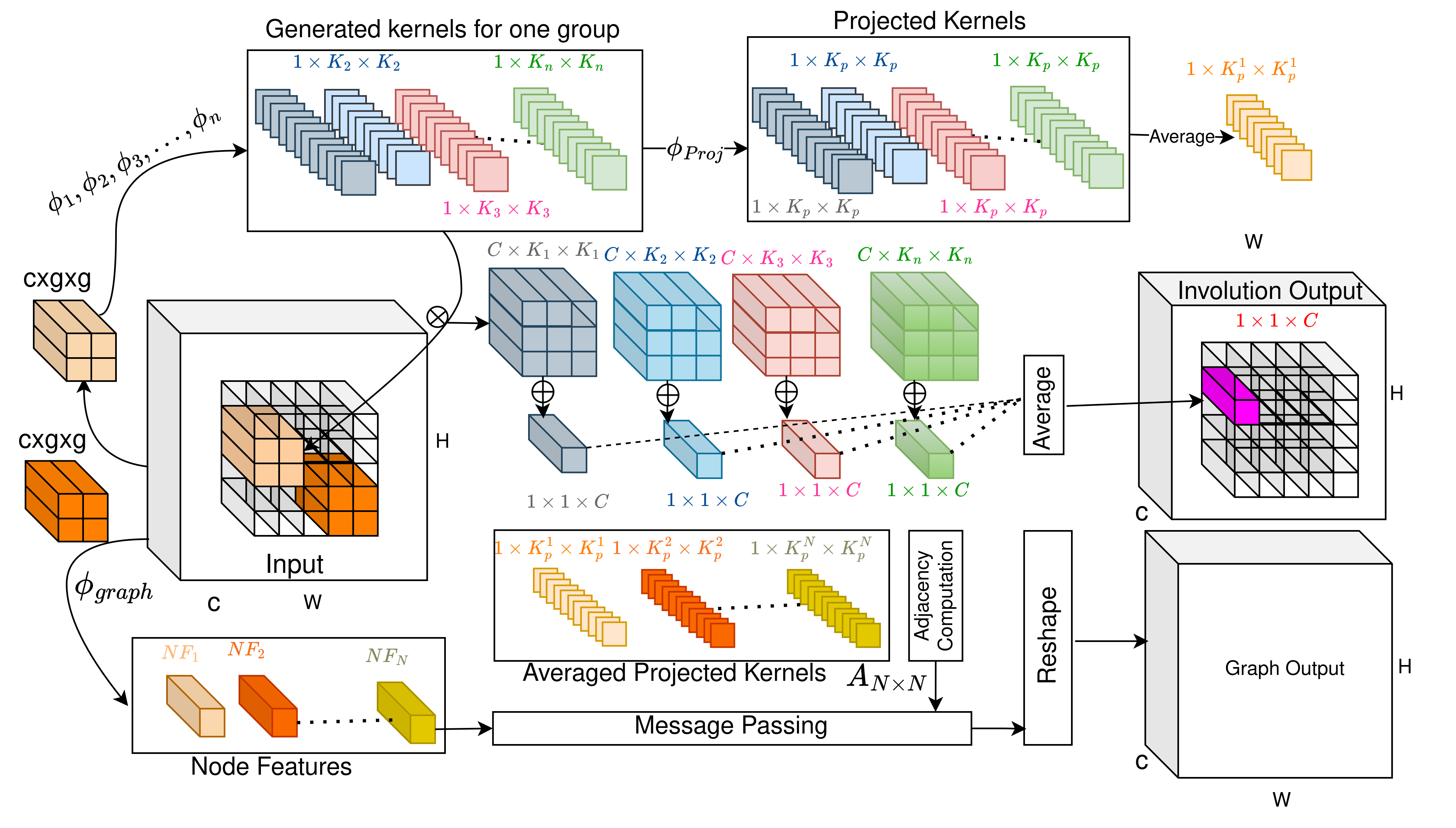}
    \caption{Illustration of the G-MSGI layer. The input tensor of shape $C \times H \times W$ is divided into spatial groups (e.g., $2 \times 2$, shown in different colors). Each group is processed by $n$ multi-scale kernel generators $\{\phi_i\}_{i=1}^n$, producing kernels that are applied across all pixels in the group (visualized here for a single pixel in pink for simplicity), and their outputs are averaged to obtain the involution result. Simultaneously, node features for each group are computed via $\phi_{\text{graph}}$, and each kernel is projected using a corresponding projection function $\{\phi_{\text{proj}_i}\}$. These projected kernels are averaged to form node descriptors (colored to indicate node identity), which are used to compute the adjacency matrix $A$ for graph construction. Graph message passing is then performed over these nodes to obtain a graph-enhanced feature representation. Circles marked with $\otimes$ and $\oplus$ denote element-wise multiplication and addition, respectively.}

    \label{fig:GMSGILayer}
\end{figure*}

The following subsections describe each core component of a G-MSGI layer in detail, beginning with the grouped pixel-level involution module. The overall flow and interaction of these components within the G-MSGI layer are visually illustrated in Figure~\ref{fig:GMSGILayer}.

\subsection{Grouped Pixel-Level Involution}
In contactless fingerprint recognition, accurately extracting minutiae depends on effectively capturing subtle local variations in ridge-valley patterns. Unlike standard convolution, involution generates adaptive kernels dynamically conditioned on local image features, providing enhanced spatial adaptivity crucial for nuanced structural representation. However, applying involution such that each pixel independently produces its own unique kernel introduces significant computational redundancy. This redundancy arises from the inherent spatial coherence within fingerprint ridge patterns, typically spanning approximately eight pixels in width. Neighboring pixels along a single ridge frequently exhibit nearly identical contextual information, differing minimally at micro-level details such as sweat pores, thus not necessitating distinct adaptive kernels for each pixel for accurate minutiae detection.

To alleviate this inefficiency while maintaining the adaptivity benefits of involution, a \textit{grouped pixel-level involution} strategy is proposed. This strategy partitions the input feature map $x \in \mathbb{R}^{B \times C \times H \times W}$, where $B$ denotes batch size, $C$ the number of channels, and $H \times W$ the spatial dimensions into non-overlapping spatial groups  of size $g \times g$. Each group is processed as a unified computational unit for adaptive kernel generation, significantly reducing redundancy while preserving essential information.

Specifically, contextual representations within each spatial group are aggregated using spatial average pooling to obtain the grouped feature map $x_g$:
\begin{equation}\label{eq:2group_pool}
    x_g = \text{AvgPool}_g(x) \in \mathbb{R}^{B \times C \times \frac{H}{g} \times \frac{W}{g}}
\end{equation}

This pooling operation effectively aggregates local features within each $g\times g$ block, capturing predominant ridge-valley patterns and filtering out pixel-level noise, thus preserving the key structural features necessary for minutiae extraction.

The kernel generation process for grouped involution can be represented by a function $\phi$ that transforms the grouped feature map into adaptive kernels. Following the principles of traditional involution, this process begins with a dimensionality reduction step:
\begin{equation}\label{eq:dimension_reduction}
    z = \text{ReLU}(\text{BN}(W_{\text{red}} * x_g)) \in \mathbb{R}^{B \times C_{\text{red}} \times \frac{H}{g} \times \frac{W}{g}}
\end{equation}

\noindent where $W_{\text{red}}$ represents the weights of a $1 \times 1$ convolution that reduces the channel dimension from $C$ to $C_{\text{red}}$. This reduction creates a compact representation that captures essential features while minimizing computational overhead. The batch normalization (BN) and ReLU activation stabilize training and introduce non-linearity to enhance the representational capacity.

The compact representation $z$ is then transformed into adaptive kernels through a second $1 \times 1$ convolution followed by a sigmoid activation:
\begin{equation}\label{eq:kernel_generation}
    \mathcal{K} = \sigma(W_{\text{span}} * z) \in \mathbb{R}^{B \times G \times K^2 \times \frac{H}{g} \times \frac{W}{g}}
\end{equation}

\noindent where $G$ is the number of channel groups, $K$ is the kernel size, and $\sigma$ represents the sigmoid activation function. The sigmoid activation constrains kernel values to $(0,1)$, ensuring stable feature transformation. This two-step process can be concisely expressed as:
\begin{equation}\label{eq:kernel_phi}
    \mathcal{K} = \phi(x_g)
\end{equation}

Similar to conventional involution approaches, channel groups are employed in this formulation. The input channels are divided into $G$ groups, with kernels generated separately for each group. This channel-wise grouping allows for specialized feature extraction across different feature subspaces. For each channel group $g \in \{1, 2, \ldots, G\}$, the corresponding subset of channels $C_g = \{C_{g-1}+1, C_{g-1}+2, \ldots, C_g\}$ where $C_g - C_{g-1} = C/G$ shares the same kernel parameters:
\begin{equation}
\mathcal{K}_g = \phi_g(x_g)
\end{equation}

While generating one shared kernel per group significantly reduces redundancy, applying these group-level kernels at individual pixels within the group remains critical. Although intra-group regions exhibit substantial structural similarity due to ridge coherence, minutiae points are sparse, highly localized, and may appear at arbitrary pixel locations. Additionally, since the overall objective extends beyond minutiae detection towards capturing detailed fingerprint features for global representation, applying adaptive kernels at every pixel ensures finer spatial details are retained. To balance efficiency with detailed spatial representation, the generated group-level kernels are therefore expanded across their corresponding $g \times g$ blocks: 
\begin{equation}\label{eq:kernel_expansion} \mathcal{K}_{\text{expanded}}(b, g, k, i, j) = \mathcal{K}(b, g, k, \lfloor i/g \rfloor, \lfloor j/g \rfloor) \in \mathbb{R}^{B \times G \times K^2 \times H \times W}, \end{equation} 

where $\lfloor \cdot \rfloor$ denotes the floor operation. This kernel expansion operation replicates the adaptive kernel across each spatial group's pixels, maintaining detailed pixel-level responsiveness within the computationally efficient grouped structue.

The final step in the grouped involution process is the application of these expanded kernels to the input feature map. The output $y_{\text{inv}}$ at each pixel location $(i,j)$ is computed by applying the expanded adaptive kernels across the $K \times K$ local neighborhoods $\mathcal{N}_K(i,j)$:
\begin{equation}\label{eq:grouped_involution}
    y_{\text{inv}}(i,j) = \sum_{k \in \mathcal{N}_K(i,j)} \mathcal{K}_{\text{expanded}}(b, g, k, i, j) \cdot x(b, c, i + \Delta k_1, j + \Delta k_2)
\end{equation}

\noindent yielding the output feature map $y_{\text{inv}} \in \mathbb{R}^{B \times C \times H \times W}$, where $\Delta k_1$ and $\Delta k_2$ represent the relative spatial offsets within the kernel neighborhood.

This grouped pixel-level involution approach not only reduces computational redundancy while maintaining spatial adaptivity but also introduces an organized spatial structure that naturally facilitates graph-based modeling, as discussed in subsequent sections. 

\subsection{Multi-Scale Dynamic Kernel Generation}
Grouped pixel-level involution enables localized adaptive filtering, but capturing fingerprint structures effectively demands a multi-scale approach. Minutiae points are fine-scale features confined to small neighborhoods, whereas ridge flow continuity extends across broader spatial regions. A fixed receptive field limits the ability to represent both granular and global structural details simultaneously. Overcoming this constraint requires a mechanism that dynamically adjusts spatial sensitivity to accommodate varying levels of information.

To this end, a multi-scale dynamic kernel generation strategy is employed. This mechanism extends the grouped involution framework by generating adaptive kernels at multiple spatial resolutions, all derived from the same grouped feature representation $x_g \in \mathbb{R}^{B \times C \times \tfrac{H}{g} \times \tfrac{W}{g}}$ as defined in Equation~\ref{eq:2group_pool}. For each kernel size $K \in \text{KS}$, where $\text{KS}$ denotes the set of predefined kernel sizes, a scale-specific transformation $\phi_K$ produces adaptive kernels:
\begin{equation}\label{eq:multiscale_kernel_transform}
\mathcal{K}^{(K)} = \phi_K(x_g) \in \mathbb{R}^{B \times G \times K^2 \times \tfrac{H}{g} \times \tfrac{W}{g}}.
\end{equation}

Each $\phi_K$ is defined by a reduction followed by an expansion operation, mirroring the grouped involution formulation but parameterized independently per scale:
\begin{equation}\label{eq:multiscale_transform_detail}
\phi_K(x_g) = \sigma(W^{(K)}_{\text{span}} * \text{ReLU}(\text{BN}(W^{(K)}_{\text{red}} * x_g))).
\end{equation}

Here, $W^{(K)}_{\text{red}} \in \mathbb{R}^{C_{\text{red}} \times C}$ and $W^{(K)}_{\text{span}} \in \mathbb{R}^{(G \cdot K^2) \times C_{\text{red}}}$ are learnable convolutional weights for scale $K$, and $\sigma$ denotes the sigmoid activation. This structure enables the network to modulate the receptive field in a scale-aware manner, adapting to both fine and coarse fingerprint structures.

To preserve spatial consistency for each pixel during kernel application, the group-level kernels are spatially broadcasted across their corresponding blocks:
\begin{equation}\label{eq:expanded_kernel}
\mathcal{K}^{(K)}_{\text{expanded}}(b, g, k, i, j) = \mathcal{K}^{(K)}(b, g, k, \lfloor i/g \rfloor, \lfloor j/g \rfloor).
\end{equation}

The scale-specific feature maps are then computed via grouped involution using the expanded kernels:
\begin{equation}\label{eq:scale_specific_output}
y^{(K)}_{\text{inv}}(i,j) = \sum_{k \in \mathcal{N}_K(i,j)} \mathcal{K}^{(K)}_{\text{expanded}}(b, g, k, i, j)\cdot x(b, c, i + \Delta k_1, j + \Delta k_2).
\end{equation}

The final multi-scale output is obtained by aggregating features across all scales:
\begin{equation}\label{eq:multi_scale_integration}
y_{\text{inv}} = \frac{1}{|\text{KS}|} \sum_{K \in \text{KS}} y^{(K)}_{\text{inv}}.
\end{equation}

This aggregation provides a rich, unified feature representation that captures both local minutiae and global ridge structures. The resulting multi-scale features support robust fingerprint encoding and provide a structured input for the graph-based modeling described in the following section.

\subsection{Graph-Based Relational Modeling}
While the previously described involution mechanisms excel at capturing localized structural patterns, fingerprint recognition ultimately depends on understanding the global arrangement of these patterns. Minutiae points derive their discriminative power not just from their local appearance but from their relative positioning within the overall fingerprint structure. To capture these long-range dependencies efficiently, a graph-based relational modeling approach is developed that builds upon the spatial grouping established during the grouped involution framework.

\begin{algorithm}[H]
\caption{Forward Pass of a G-MSGI Layer}
\label{alg:gmsgi}
\small
\begin{algorithmic}[1]
\Require Input feature map $x \in \mathbb{R}^{B \times C_{\text{in}} \times H \times W}$
\Ensure Updated feature map $y \in \mathbb{R}^{B \times C \times H \times W}$
\Statex \parbox{\dimexpr\linewidth-\algorithmicindent\relax}{ \textbf{Parameters:} $C$: output channels, $g$: group size, $\text{KS}$: kernel sizes, $G$: channel groups, $s$: stride, $r$: reduction ratio}

\State Align input channels via $1 \times 1$ convolution if $C_{\text{in}} \ne C$
\State Downsample input if stride $s > 1$
\State Compute grouped features: $x_g \gets \text{AvgPool}_g(x)$
\ForAll{$K \in \text{KS}$}
    \State Generate adaptive kernel $\mathcal{K}^{(K)} \gets \phi_K(x_g)$
    \State Expand $\mathcal{K}^{(K)}$ across spatial groups
    \State Apply grouped involution with $\mathcal{K}^{(K)}$ to get $y^{(K)}_{\text{inv}}$
\EndFor
\State Aggregate multi-scale outputs from involution: $y_{\text{inv}} \gets \frac{1}{|\text{KS}|} \sum_{K \in \text{KS}} y^{(K)}_{\text{inv}}$

\State Project $x$ to node features using $1 \times 1$ convolution
\State Unfold node features into grouped patches
\ForAll{$K \in \text{KS}$}
    \State Project $\mathcal{K}^{(K)}$ to latent space: $\alpha^{(K)} \gets \varphi_{\text{proj}}^{(K)}(\mathcal{K}^{(K)})$
\EndFor
\State Compute mean descriptor: $\bar{\alpha} \gets \frac{1}{|\text{KS}|} \sum_{K \in \text{KS}} \alpha^{(K)}$
\State Compute adjacency matrix $A$ via scaled cosine similarity over $\bar{\alpha}$
\State Perform message passing: $x'_i \gets \text{ReLU}(\sum_j A_{ij} \cdot x_j)$
\State Fold updated node features to obtain $y_{\text{graph}}$

\State Combine outputs: $y \gets y_{\text{inv}} + \mu \cdot y_{\text{graph}}$
\State \Return{$y$}
\end{algorithmic}
\end{algorithm}

Let $x \in \mathbb{R}^{B \times C \times H \times W}$ denote the feature map. A shared $1 \times 1$ convolution transforms $x$ into $x_{\text{graph}} \in \mathbb{R}^{B \times C \times H \times W}$, projecting features into a representation space more suitable for graph-based reasoning. The spatial domain is partitioned into $N = \frac{H}{g} \cdot \frac{W}{g}$ non-overlapping groups of size $g \times g$, with each group flattened to form a node embedding:
\begin{equation}
x_{\text{node}} \in \mathbb{R}^{B \times N \times (C \cdot g^2)}.
\end{equation}

Graph edges reflecting structural similarities are defined using the multi-scale involution kernels $\mathcal{K}^{(K)} \in \mathbb{R}^{B \times G \times K^2 \times \tfrac{H}{g} \times \tfrac{W}{g}}$ from the previous module. These kernels encode structural information through learned local transformations and are projected into a shared latent space using scale-specific functions:
\begin{equation}
\alpha^{(K)} = \varphi_{\text{proj}}^{(K)}(\mathcal{K}^{(K)}) \in \mathbb{R}^{B \times G \times p^2 \times \tfrac{H}{g} \times \tfrac{W}{g}}.
\end{equation}

Projections across all kernel scales are averaged to yield a unified kernel representation:
\begin{equation}
\bar{\alpha} = \frac{1}{|\text{KS}|} \sum_{K \in \text{KS}} \alpha^{(K)} \in \mathbb{R}^{B \times G \times p^2 \times \tfrac{H}{g} \times \tfrac{W}{g}}.
\end{equation}

Each $\bar{\alpha}^{(g)}$ is flattened across spatial dimensions to form node-wise kernel descriptors $\alpha_i^{(g)} \in \mathbb{R}^d$, where $d = p^2$. These kernel-derived descriptors encode structural information more robustly than raw node features, as nodes with similar local structures produce similar kernels. This enables capturing structural similarities between non-adjacent fingerprint regions with similar ridge flow patterns.

For each channel group $g \in \{1, \dots, G\}$, a separate adjacency matrix $A^{(g)} \in \mathbb{R}^{B \times N \times N}$ is constructed using scaled cosine similarity followed by softmax normalization:
\begin{equation}
A_{ij}^{(g)} = \frac{\exp\left(\frac{\alpha_i^{(g)} \cdot \alpha_j^{(g)}}{\sqrt{d}}\right)}{\sum_k \exp\left(\frac{\alpha_i^{(g)} \cdot \alpha_k^{(g)}}{\sqrt{d}}\right)}.
\end{equation}

The scaling factor $\sqrt{d}$ prevents dot products from growing too large in magnitude as dimension increases, ensuring each node maintains connections with multiple structurally similar nodes. The final adjacency matrix $A \in \mathbb{R}^{B \times N \times N}$ is computed by averaging across channel groups:
\begin{equation}
A = \frac{1}{G} \sum_{g=1}^{G} A^{(g)}.
\end{equation}

Message passing aggregates information across structurally related nodes, with updated node features $x' \in \mathbb{R}^{B \times N \times (C \cdot g^2)}$ computed as:
\begin{equation}
x'_i = \text{ReLU}\left(\sum_j A_{ij} \cdot x_j\right).
\end{equation}

This non-linear transformation captures complex relationships between structurally similar regions, with ReLU activation introducing sparsity to focus on the most relevant structural connections. The updated embeddings are folded back into the original spatial format:
\begin{equation}
y_{\text{graph}} \in \mathbb{R}^{B \times C \times H \times W}.
\end{equation}

A residual fusion integrates global structural reasoning with local adaptivity:
\begin{equation}
y_{\text{graph}} = y_{\text{graph}} + \mu \cdot (y_{\text{inv}} - y_{\text{graph}}),
\end{equation}
where $\mu$ is a learnable parameter that balances the contribution of local and global features based on specific fingerprint image characteristics.

Through this formulation, each spatial group contributes not only localized minutiae-sensitive information but also participates in a learned global fingerprint structure, captured via kernel-guided graph reasoning. This synergy between local adaptivity and non-local contextualization serves as the foundation for the full G-MSGINet architecture, which integrates these components into a unified processing pipeline described in the following section. The step-by-step forward computation within a single G-MSGI layer is summarized in Algorithm~\ref{alg:gmsgi}.

\subsection{Overall Network Architecture}

\begin{figure*}
    \centering
    \includegraphics[width=0.9\linewidth]{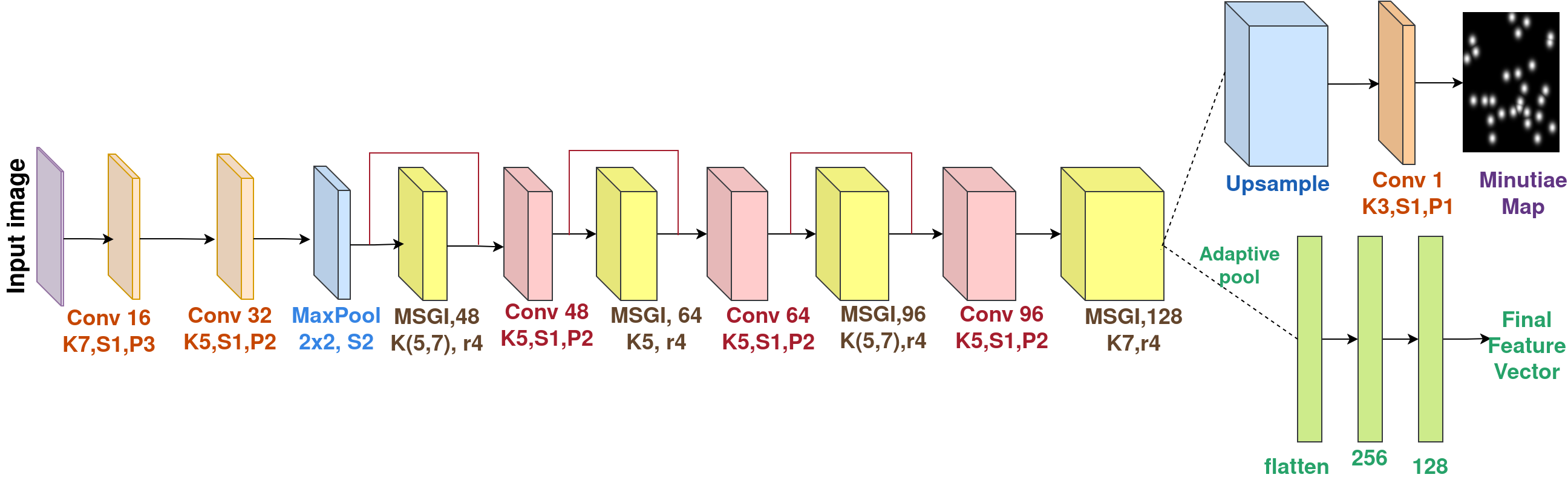}
    \caption{G-MSGINet architecture. The input image is processed through convolutional and MSGI layers (in yellow), producing two outputs: a $400 \times 400$ minutiae heatmap and a 128-dimensional feature vector. }
    \label{fig:architecture}
\end{figure*}

Building upon the adaptive modeling capabilities of the G-MSGI layer, the complete fingerprint recognition network as illustrated in Fig.~\ref{fig:architecture} is designed to jointly extract spatially precise minutiae features and globally structured identity embeddings from full fingerprint images. The architecture integrates multiple G-MSGI layers in a stacked configuration, allowing local and relational representations to evolve across the network depth.

The network begins with a downsampling phase composed of two convolutional layers with kernel sizes of \(7 \times 7\) and \(5 \times 5\), each followed by batch normalization and ReLU activation, and a subsequent max-pooling layer with a \(2 \times 2\) kernel and stride 2. Together, this phase reduces the spatial resolution of the input fingerprint image, effectively lowering computational overhead while retaining essential structural information for further processing.

The resulting feature map from this downsampling phase serves as the input to the main processing backbone, which consists of a sequence of four G-MSGI layers. Each layer incrementally refines the feature representation while preserving both spatial and structural consistency. Between these layers, lightweight convolutional refinement blocks equipped with batch normalization and ReLU activation are inserted to enhance feature expressiveness and promote stable training. Residual connections are used throughout to facilitate gradient flow and enable effective feature reuse. At each stage, the output from one layer seamlessly transitions as input to the next, ensuring progressive enhancement of fingerprint-specific representations across the network depth.

In the final transformation phase, the refined feature map is processed through two lightweight and parallel heads. Unlike existing approaches that rely on complex decoder structures, our method uses a single \(3 \times 3\) convolution to upsample the feature map to the original input resolution, producing a high-resolution \textit{minutiae likelihood map} of size \(1 \times 400 \times 400\). In parallel, a compact identity embedding is generated by applying global average pooling followed by just two fully connected layers, yielding a 256-dimensional \textit{minutiae-aware feature vector}. This minimalistic yet effective design enables joint optimization of both minutiae localization and identity representation without introducing significant overhead.

\subsection{Loss Function}

The proposed framework is trained using a composite loss that jointly supervises minutiae localization and identity embedding. These two tasks are optimized using a segmentation loss and a metric learning loss, respectively.

\paragraph{Minutiae Loss}
To address the severe class imbalance between sparse minutiae points and the dominant background pixels, the Focal Tversky Loss \cite{tversky} is employed. This loss assigns asymmetric penalties to false positives and false negatives and emphasizes harder examples using a focusing parameter $\gamma$. It promotes sharper, more confident predictions at minutiae locations.

Let \( M \) denote the binary ground truth minutiae map and \( \hat{M}\) denote the predicted confidence map. The minutiae loss is computed as:
\[
\mathcal{L}_{\text{min}} = \left(1 - \frac{2 \sum \hat{M} \cdot M}{2 \sum \hat{M} \cdot M + \alpha \sum \hat{M} \cdot (1 - M) + \beta \sum M \cdot (1 - \hat{M}) + \epsilon} \right)^{\gamma}
\]
where \(\alpha\) and \(\beta\) control the penalty for false positives and false negatives respectively, and \(\epsilon\) ensures numerical stability.

\paragraph{Feature Loss}
For learning identity embeddings, a simplified version of the SubCenter MagFace Loss \cite{meng2021magface} is used. It compares normalized feature vectors to class centers with an angular margin, improving class separation. A single sub-center per class is used, and the final prediction is optimized via cross-entropy loss. This yields the identity embedding loss denoted by $\mathcal{L}_{\text{feat}}$.

\paragraph{Composite Loss.}
To balance both objectives, an adaptive weighting strategy is applied based on the relative magnitudes of the individual losses. The final composite loss is defined as:
\[
\mathcal{L}_{\text{total}} = 2w_{\text{min}} \cdot \mathcal{L}_{\text{min}} + w_{\text{feat}} \cdot \mathcal{L}_{\text{feat}},
\]
where the weights $w_{\text{min}}$ and $w_{\text{feat}}$ are computed adaptively during training:
\[
w_{\text{min}} = \frac{1}{\frac{\mathcal{L}_{\text{min}}}{\mathcal{L}_{\text{min}} + \mathcal{L}_{\text{feat}} + \epsilon} + \epsilon}, \quad
w_{\text{feat}} = \frac{1}{\frac{\mathcal{L}_{\text{feat}}}{\mathcal{L}_{\text{min}} + \mathcal{L}_{\text{feat}} + \epsilon} + \epsilon}.
\]
This formulation ensures the network effectively learns both localization and identity discrimination without requiring orientation labels or clustering heuristics.

\section{Experiments and Experimental Results}\label{experiments}
This section presents an extensive evaluation of the proposed G-MSGINet architecture on three widely used contactless fingerprint datasets. Since G-MSGINet jointly produces both a minutiae likelihood map and a global fingerprint embedding, the evaluation is structured to comprehensively assess both local and global performance aspects. The following experimental objectives are defined to validate the effectiveness of the proposed method across various perspectives:
\begin{itemize}
\setlength{\itemsep}{0pt}
	\setlength{\parskip}{0pt} 
	\setlength{\parsep}{0pt}
    \item Evaluate minutiae detection accuracy using the generated heatmaps across three contactless fingerprint datasets.
    \item Analyze the discriminative capability of the learned feature vectors through intra-class and inter-class distance metrics.
    \item Assess fingerprint recognition performance using the extracted global feature embeddings.
    \item Compare the computational footprint and parameter count of G-MSGINet with conventional CNN and graph-based baselines.

\end{itemize}
Details of the datasets, evaluation protocols, and implementation settings are provided in the subsequent subsections.

\subsection{Dataset}
The evaluation is conducted using three publicly available contactless fingerprint datasets: PolyU\cite{lin2018matching}, Benchmark 2D/3D\cite{zhou2014benchmark}, and CFPose\cite{tan2020towards}. The PolyU Cross dataset includes two sessions, the first contains 2,016 images from 336 fingers (six impressions per finger), while the second provides 960 images from 160 of the same fingers, collected with a time interval ranging from 2 to 24 months. The Benchmark 2D/3D dataset comprises 9,000 fingerprint images from 1,500 fingers, captured across three different viewpoints with two impressions per view. Since ground truth minutiae annotations are not available for these datasets, a subset of 100 unique fingers was randomly selected from both PolyU and Benchmark, and their minutiae points were manually annotated, resulting in 1,200 and 200 labeled images, respectively. The CFPose dataset includes 20 fingerprint images with 2 impresssions and 100 fingerprint images with the only one impressions, in total it has 140 individuals, with publicly available minutiae annotations. All the annotated minutiae information is made publicly available to support further research in this area.

For training, the first session of the PolyU dataset serves as the primary source. Due to the limited number of impressions in the Benchmark and CFPose datasets, data augmentation techniques, such as rotation, noise addition, scaling, translation, and blurring are applied to generate 6 impressions per finger. From these, 3 impressions are used for training and the remaining 3 for testing and recognition evaluation, consistent with the protocol followed for PolyU. The model is initially trained on the PolyU dataset and subsequently fine-tuned on the Benchmark and CFPose datasets.

\subsection{Experimental Environment}

All implementation, training, and evaluation procedures were carried out using Python v3.10.13 within a conda-managed virtual environment (conda v23.10.0). The primary libraries used include PyTorch (v2.1.0) for deep learning, NumPy (v1.24.3) and SciPy (v1.10.1) for numerical operations, Pandas (v2.0.2) for data handling, scikit-learn (v1.2.2) for metric computation, Matplotlib (v3.7.1) for visualization, and OpenCV (v4.7.0.72) for image processing. All experiments were conducted on a Linux-based workstation running Ubuntu 22.04 LTS, equipped with an Intel(R) Core(TM) i9 processor, 128GB RAM, a 2TB NVMe SSD, and an NVIDIA GeForce RTX 4090 GPU (24GB VRAM) with CUDA v12.4 and driver version 550.127.08.

\subsection{Implementation Details}
The model training employed the Adam optimizer with a weight decay of \(1 \times 10^{-8}\), running for 200 epochs with a batch size of 4. The intermediate CNN layers incorporated a dropout rate of 0.02 to mitigate overfitting. The G-MSGI layers were configured with a stride of 1, a group size of 8, a reduction ratio of 4, a dilation value of 1, and projected kernel sizes of 5. These settings ensured efficient feature extraction and model optimization.

\subsection{Minutiae Extraction Results}

To train the model for minutiae localization, the ground truth annotations are prepared following the protocol used by ContactlessMinutNet~\cite{zhang2021multi}. Specifically, a Gaussian heatmap is generated for each ground truth minutiae point and used as a supervision signal. During inference, the predicted minutiae map is post-processed using Non-Maximum Suppression (NMS) with a threshold of 0.95 to extract final minutiae locations from the response map.

\begin{figure}[!ht]
    \centering
    \includegraphics[width=0.9\linewidth]{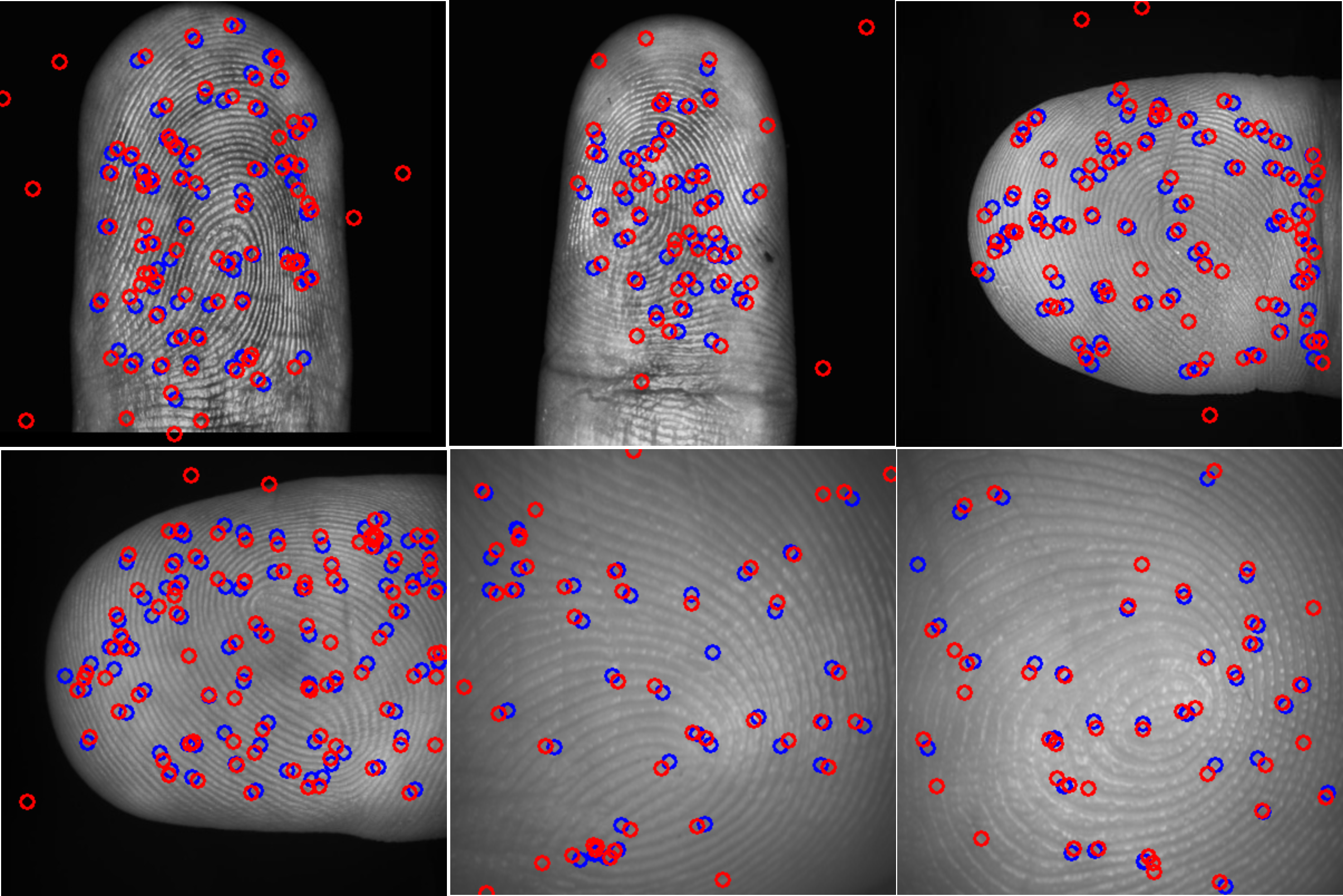}
    \caption{Predicted minutiae points (red circles) overlaid on sample images from CFPose, Benchmark 2D/3D, and PolyU datasets (top to bottom, left to right). }
    \label{fig:minutiaepoints}
\end{figure}

Figure \ref{fig:minutiaepoints} displays representative visualizations of predicted minutiae locations across three datasets: CFPose, Benchmark 2D/3D, and PolyU. For each dataset, two sample fingerprint images are shown with predicted minutiae points indicated by red circles and ground truth minutiae marked with blue circles. These visualizations demonstrate G-MSGINet's ability to accurately identify minutiae points across diverse fingerprint patterns and image qualities.

The performance of the proposed G-MSGINet for minutiae detection is evaluated using standard metrics: precision, recall, and F1 score. A predicted minutiae is considered correct if it lies within a specified pixel distance from its corresponding ground-truth location. This spatial tolerance, denoted as “px” in the result tables, defines the maximum allowable distance between a predicted and actual minutia point for the prediction to be counted as correct. A threshold of 8 pixels is used for G-MSGINet throughout the experiments, reflecting a strict yet commonly adopted evaluation setting in fingerprint recognition, roughly equivalent to one ridge width. For comparative methods, the same thresholds used in their original evaluations are retained and explicitly shown in the “px” column for clarity and fairness.

\begin{table}[H]
\centering
\caption{Minutiae detection results on the CFPose dataset.}
\label{tab:cfpose_mresults}
\renewcommand{\arraystretch}{0.9}
\begin{tabular}{lcccc}
\toprule
Method & px & Precision & Recall & F1 \\
\midrule
MINDTCT~\cite{ko2007user} & 8 & 0.195 & 0.129 & 0.151 \\
Verifinger~\cite{neurotechnology2025} & 8 & 0.338 & 0.393 & 0.359 \\
ContactlessNet~\cite{tan2020towards} & 8 & 0.733 & 0.759 & 0.741 \\
ContactlessNet~\cite{tan2020towards} & 16 & 0.764 & 0.790 & 0.772 \\
WSMS-CNet~\cite{cotrim2022multiscale} & 10 & 0.724 & 0.899 & 0.820 \\
ReUSE-CNet~\cite{cotrim2023residual} & 10 & 0.749 & 0.912 & 0.823 \\
G-MSGINet (Proposed) & 8 & \textbf{0.811} & \textbf{0.917} & \textbf{0.860} \\
\bottomrule
\end{tabular}
\end{table}

Results are presented individually for each benchmark dataset (CFPose, Benchmark 2D/3D, and PolyU) in separate tables. This separate presentation is necessary because the baseline methods were originally proposed, validated, and evaluated independently on specific datasets under varying conditions and thresholds.

On the CFPose dataset (Table~\ref{tab:cfpose_mresults}), G-MSGINet demonstrates clear superiority by achieving a balanced combination of high precision and recall. It outperforms previous deep learning approaches such as WSMS-CNet and ReUSE-CNet, even though these methods operate with more lenient pixel thresholds.

\begin{table}[H]
\centering
\caption{Minutiae detection results on the Benchmark 2D/3D dataset.}
\label{tab:benchmark_mresults}
\renewcommand{\arraystretch}{0.9}
\begin{tabular}{lcccc}
\toprule
Method & px & Precision & Recall & F1 \\
\midrule
MINDTCT\cite{ko2007user} & 8 & 0.144 & 0.070 & 0.093 \\
Verifinger\cite{neurotechnology2025} & 8 & 0.332 & 0.502 & 0.395 \\
ContactlessNet~\cite{tan2020towards} & 8 & 0.701 & 0.654 & 0.672 \\
ContactlessNet~\cite{tan2020towards} & 16 & 0.704 & 0.640 & 0.665 \\
ContactlessMinutNet~\cite{zhang2021multi} & 12 & 0.791 & 0.809 & 0.800 \\
WSMS-CNet~\cite{cotrim2022multiscale} & 10 & 0.726 & 0.829 & 0.775 \\
ReUSE-CNet~\cite{cotrim2023residual} & 10 & 0.714 & 0.868 & 0.784 \\
G-MSGINet (Proposed) & 8 & \textbf{0.893} & \textbf{0.860} & \textbf{0.876} \\
\bottomrule
\end{tabular}
\end{table}

Similarly, results on the Benchmark 2D/3D dataset (Table~\ref{tab:benchmark_mresults}) confirm that G-MSGINet maintains superior overall performance, exhibiting robustness to variations in fingerprint patterns and quality. Despite using the stricter 8-pixel threshold, it surpasses methods previously evaluated with relaxed tolerances.

On the PolyU dataset (Table~\ref{tab:polyu_mresults}), the proposed method again leads in all three evaluation metrics. Despite fewer published benchmarks for this dataset, ContactlessNet remains a strong baseline. G-MSGINet exceeds its performance even at the same threshold, reinforcing the model’s ability to generalize across datasets with varying image properties.

\begin{table}[H]
\centering
\caption{Minutiae detection results on the PolyU dataset.}
\label{tab:polyu_mresults}
\renewcommand{\arraystretch}{0.9}
\begin{tabular}{lcccc}
\toprule
Method & px & Precision & Recall & F1 \\
\midrule
MINDTCT~\cite{ko2007user} & 8 & 0.100 & 0.124 & 0.106 \\
Verifinger~\cite{neurotechnology2025} & 8 & 0.425 & 0.353 & 0.381 \\
ContactlessNet~\cite{tan2020towards} & 8 & 0.604 & 0.664 & 0.628 \\
ContactlessNet~\cite{tan2020towards} & 16 & 0.704 & 0.640 & 0.665 \\
G-MSGINet (Proposed) & 8 & \textbf{0.851} & \textbf{0.812} & \textbf{0.833} \\
\bottomrule
\end{tabular}
\end{table}

Figure~\ref{fig:f1_comparison} further summarizes the overall trends across the three datasets visually, clearly highlighting the consistent advantage of G-MSGINet. This visual summary emphasizes that the model’s consistently superior F1-scores are due not only to a strong balance between precision and recall but also to robustness under strict evaluation conditions

In summary, across all benchmark datasets, the proposed G-MSGINet achieves the best overall minutiae detection performance. The integration of adaptive kernel learning and graph-based relational modeling significantly enhances spatial accuracy, structural consistency, and robustness, allowing it to consistently outperform existing methods even under stringent matching criteria.

\subsection{Feature Discrimination Analysis}

In addition to predicting minutiae locations, the proposed model generates a minutiae-aware feature vector for each fingerprint, designed to encode discriminative identity-related information. To evaluate the effectiveness of these feature representations, a feature discrimination analysis is conducted using cosine similarity between pairs of normalized feature vectors. This analysis is exclusively performed on the test set, ensuring unbiased evaluation on unseen data without influence from the training process.

\begin{figure}[t]
    \centering

    \begin{subfigure}{0.49\linewidth}
        \includegraphics[width=\linewidth]{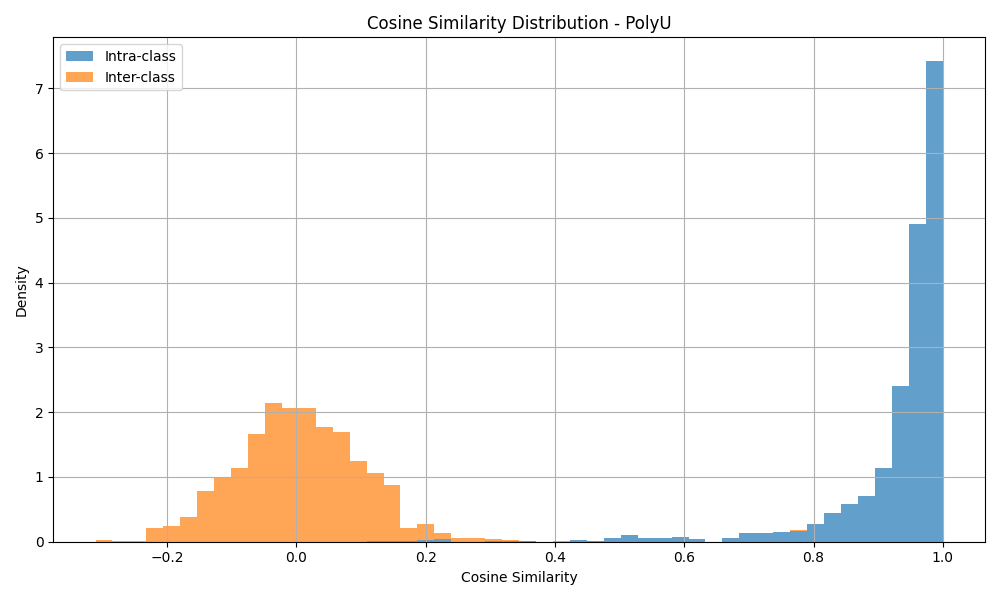}
        \caption{PolyU: Clear separation between intra- and inter-class similarities.}
        \label{fig:polyu_sim}
    \end{subfigure}
    \hfill
    \begin{subfigure}{0.49\linewidth}
        \includegraphics[width=\linewidth]{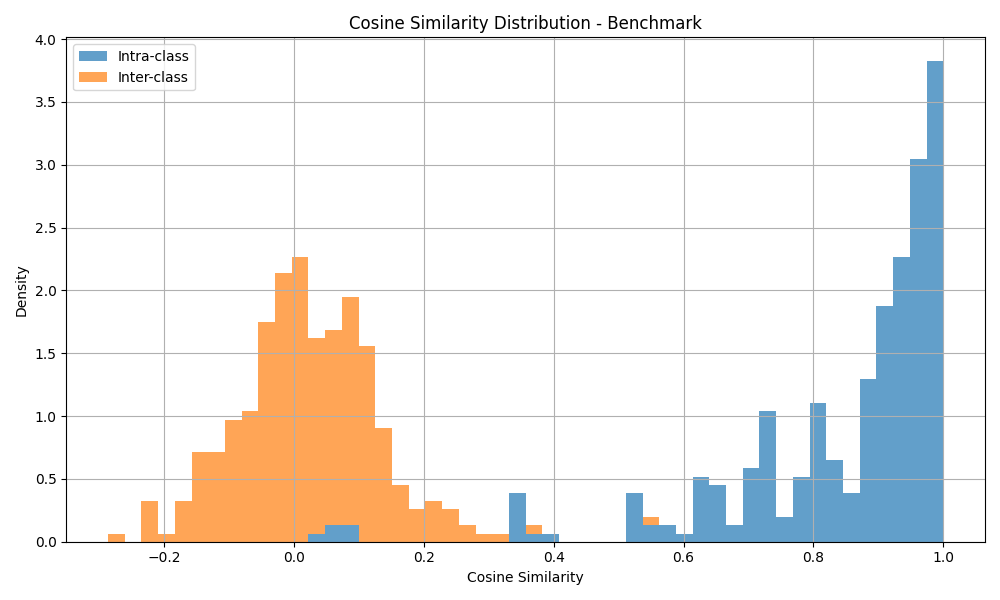}
        \caption{Benchmark: Overlap due to diverse impressions and views.}
        \label{fig:benchmark_sim}
    \end{subfigure}

    \vspace{0.5em}
    \begin{subfigure}{0.49\linewidth}
        \includegraphics[width=\linewidth]{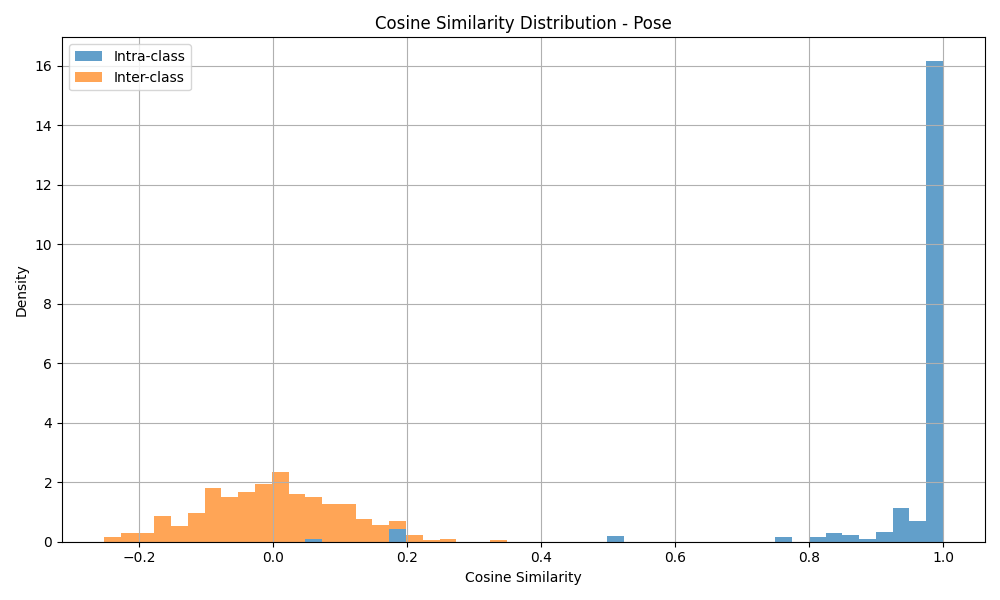}
        \caption{CFPose: Minor overlap from pose variation and data limits.}
        \label{fig:cfpose_sim}
    \end{subfigure}
    \hfill
    \begin{subfigure}{0.49\linewidth}
        \includegraphics[width=\linewidth]{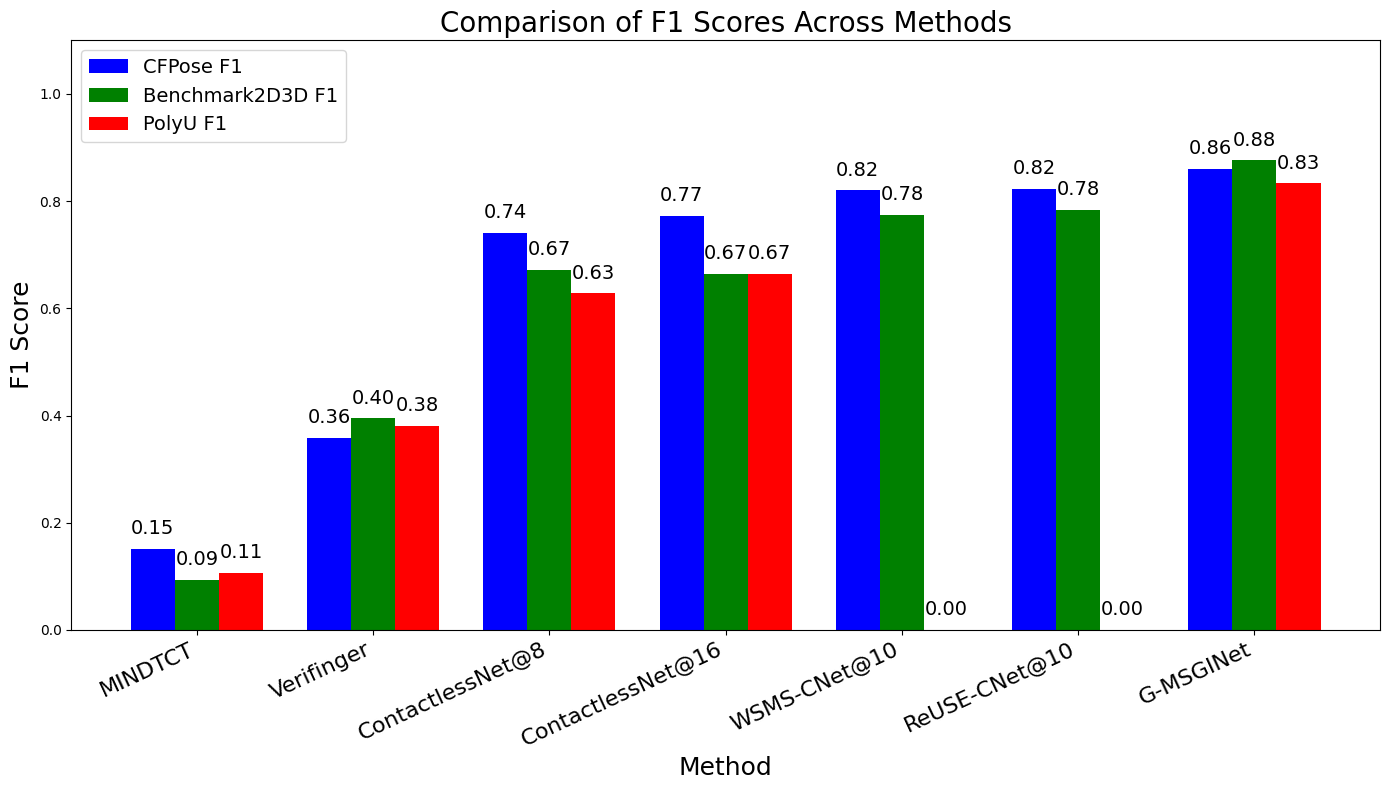}
        \caption{F1-score: G-MSGINet outperforms on all three datasets.}
        \label{fig:f1_comparison}
    \end{subfigure}

    \caption{Evaluation across three datasets: Cosine similarity distributions highlight intra- vs. inter-class feature separability, while F1-score comparisons demonstrate the superior discriminative performance of G-MSGINet.}
    \label{fig:sim_f1_analysis}
\end{figure}

For each dataset, intra-class similarities are computed by comparing all impressions of the same finger, while an equal number of inter-class pairs are randomly sampled from impressions of different fingers. These similarity distributions are then visualized using density plots, where the x-axis represents cosine similarity values and the y-axis indicates the density of pairwise similarity scores.

As illustrated in Figure~\ref{fig:polyu_sim}, the PolyU dataset exhibits a strong separation between intra- and inter-class distributions. Intra-class pairs are densely concentrated near a cosine similarity of 1, indicating that impressions of the same fingerprint generate highly similar feature vectors. In contrast, inter-class pairs are spread across lower similarity values, demonstrating the model's ability to distinguish fingerprints effectively.

Both CFPose and Benchmark datasets exhibit some level of overlap between intra- and inter-class distributions, though the degree varies, as seen in Figures~\ref{fig:cfpose_sim} and~\ref{fig:benchmark_sim}. In CFPose (Figure~\ref{fig:cfpose_sim}), the overlap is slight and primarily influenced by pose variability and a limited sample size, leading to minor inconsistencies in same-finger impressions. Despite this, the model successfully maintains meaningful feature separation, ensuring reliable discrimination across different fingerprint identities.

The Benchmark dataset (Figure~\ref{fig:benchmark_sim}) presents a more noticeable overlap between intra- and inter-class pairs. This is likely due to its diverse fingerprint impressions, multi-view acquisitions, and variations in quality. However, even with this overlap, the model demonstrates strong separability, effectively distinguishing a substantial number of fingerprint features. The total overlap is not significant enough to compromise the model’s reliability, as the majority of samples remain clearly differentiated, reinforcing the robustness of the learned representations for fingerprint identity verification.

Overall, these density plots confirm that the minutiae-aware features successfully encode identity-related information, ensuring high intra-class consistency and clear inter-class distinction across all datasets, even in challenging conditions.

\subsection{Recognition Results}
The proposed fingerprint recognition method, G-MSGINet, is comprehensively evaluated through two primary biometric tasks: \textit{verification} and \textit{closed-set identification}. Evaluations are conducted exclusively on test data to ensure impartiality and robustness in the reported performance.

\paragraph{Verification Task.}
In the verification scenario, the task involves assessing whether two fingerprint impressions originate from the same finger. Genuine pairs are created by comparing different impressions of the same finger, whereas imposter pairs are generated by matching the first impression of each finger with the first impressions of all other fingers in the dataset. Matching between pairs is quantified using the cosine similarity between the extracted minutiae-aware feature vectors.

\begin{figure}[!ht]
    \centering
    \begin{subfigure}{0.49\linewidth}
        \includegraphics[width=\linewidth]{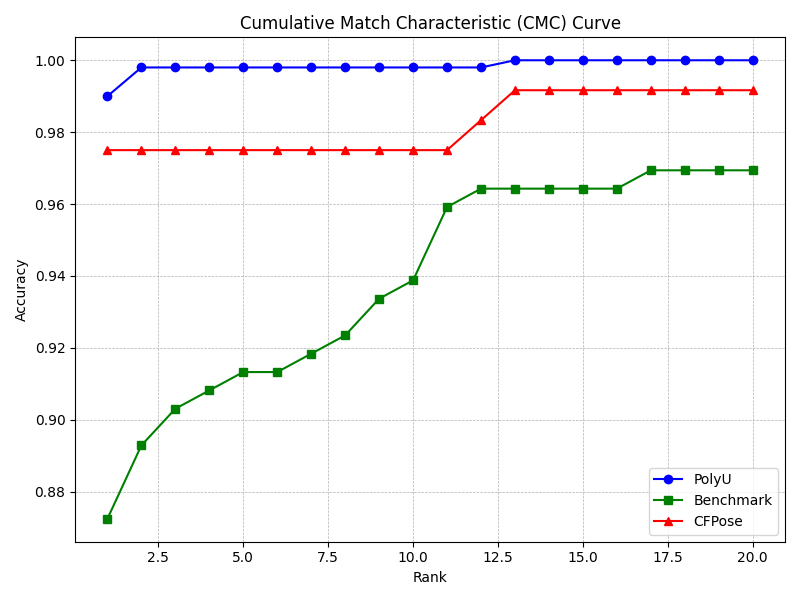}
        \caption{CMC curves showing the identification accuracy across rank positions for each dataset.}
        \label{fig:cmc_plots}
    \end{subfigure}
    \hfill
    \begin{subfigure}{0.49\linewidth}
        \includegraphics[width=\linewidth]{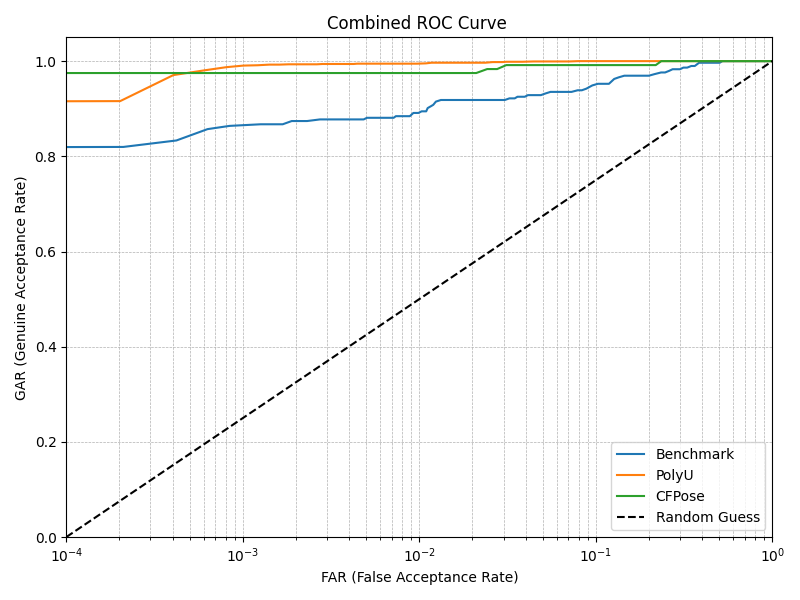}
        \caption{ROC curves illustrating the verification performance of the proposed model based on genuine acceptance across varying false acceptance rates.}
        \label{fig:roc_plots}
    \end{subfigure}
\end{figure}

In the PolyU dataset, which contains six impressions per finger for 100 fingers, a total of \(100 \times \binom{6}{2} = 1500\) genuine pairs and \(100 \times 99 = 9900\) imposter pairs are generated. The Benchmark 2D/3D dataset consists of three impressions per finger, yielding 300 genuine and 9900 imposter pairs. The CFPose dataset includes three impressions per finger for 140 fingers, resulting in 420 genuine and 19,460 imposter pairs. ROC curves generated from similarity scores are shown in Figure~\ref{fig:roc_plots}. The verification performance is quantified using standard metrics, including the Equal Error Rate (EER), Genuine Acceptance Rate (GAR), False Acceptance Rate (FAR), and the Area Under the Curve (AUC).

\paragraph{Closed-Set Identification Task.}
In the identification task, the goal is to correctly identify the query fingerprint from a pre-enrolled gallery. The first impression of each finger is enrolled as a gallery template, and subsequent impressions serve as query samples. Matching scores are computed using cosine similarity, and results are ranked accordingly. Performance is analyzed using Cumulative Matching Characteristic (CMC) curves, presented in Figure~\ref{fig:cmc_plots}, indicating identification accuracy across increasing rank positions.

\begin{table}[!ht]
    \centering
    \caption{Recognition performance of the G-MSGINet across datasets.}
    \label{tab:recognition_results}
    \small
    \begin{tabularx}{0.8\textwidth}{lXXXX}
        \toprule
        \textbf{Dataset} & \textbf{EER (\%)} & \textbf{AUC (\%)} & \textbf{Rank-1 (\%)} & \textbf{Rank-5 (\%)} \\
        \midrule
        Benchmark   & 6.0  & 0.89 & 87.2 & 91.8 \\
        PolyU & 0.5  & 0.99 & 99.0 & 100.0 \\
        CFPose      & 2.0  & 0.97 & 97.0 & 99.1 \\
        \bottomrule
    \end{tabularx}
\end{table}
The Table~\ref{tab:recognition_results} summarizes verification and identification results across three datasets, highlighting performance variations due to complexity and acquisition challenges. G-MSGINet excels on the PolyU dataset (0.5\% EER, 99.0\% Rank-1) with minimal interference, remains strong on CFPose (2.0\% EER, 97.0\% Rank-1) despite pose variations, and achieves respectable performance on the Benchmark dataset (6.0\% EER, 87.2\% Rank-1) despite viewpoint changes and background interference. These results demonstrate the robustness of the proposed minutiae-aware representations across diverse conditions.

\subsubsection{Comparison with Existing Approaches}

To validate the effectiveness of G-MSGINet, a comparative analysis against state-of-the-art approaches is conducted across all three datasets. The comparison focuses on key performance metrics: EER for verification and Rank-1 accuracy for identification. The distinct characteristics of each dataset create a progressive challenge spectrum: PolyU contains clean fingerprint images without background interference and minimal pose variations; CFPose includes significant pose variations with background regions; and Benchmark features multiple viewpoint variations with non-fingerprint background regions, presenting the most challenging recognition scenario.

On the Benchmark 2D/3D dataset (Table~\ref{tab:benchmark_results}), G-MSGINet achieves an EER of 6.0\% and a Rank-1 accuracy of 91.8\%. This performance substantially outperforms ContactlessNet (31.82\% EER, 49.86\% Rank-1) and traditional methods like RTPS+DCM (19.81\% EER, 36.25\% Rank-1). While WSMS-CNet and ReUSE-CNet show competitive EER values at 4.14\% and 4.45\% respectively, they do not report Rank-1 accuracy. It's worth noting that these methods rely heavily on minutiae orientation information and preprocessing steps, which G-MSGINet intentionally avoids. Despite this simplified approach, G-MSGINet achieves the highest Rank-1 accuracy (91.8\%) among all methods, even surpassing ContactlessMinutNet (89.60\%), demonstrating robust performance under challenging viewpoint variations.

\begin{table}[H]
\centering
\caption{Comparison of fingerprint recognition methods on the Benchmark dataset.}
\label{tab:benchmark_results}
\begin{tabularx}{0.75\textwidth}{p{4cm}XX}
\toprule
\textbf{Method} & \textbf{EER (\%)} & \textbf{Rank-1 (\%)} \\
\midrule
ContactlessNet\cite{tan2020towards} & 31.82 & 49.86 \\
WSMS-CNet\cite{cotrim2022multiscale} & 4.14 & -- \\
ReUSE-CNet\cite{cotrim2023residual} & 4.45 & -- \\
ContactlessMinutNet\cite{zhang2021multi} & 4.28 & 89.60 \\
MultisimeseCNN~\cite{lin2018cnn} & 7.11 & 58.87 \\
RTPS+DCM~\cite{lin2018matching} & 19.81 & 36.25 \\
C2CL~\cite{grosz2021c2cl} & 6.81 & -- \\
SMFI\cite{dong2023synthesis} & 13.84 & 56.7 \\
\textbf{G-MSGINet (Proposed)} & \textbf{6.0} & \textbf{91.8} \\
\bottomrule
\end{tabularx}
\end{table}

For the CFPose dataset (Table~\ref{tab:cfpose_results}), G-MSGINet delivers a strong EER of 2.0\% and achieves the highest Rank-1 accuracy (97.0\%) among all compared methods. While SMFI achieves the lowest EER at 1.24\%, it's important to highlight that SMFI relies on synthetic image generation, whereas G-MSGINet operates directly on raw captured images without such data augmentation strategies. Despite this methodological difference, G-MSGINet's superior Rank-1 accuracy demonstrates exceptional discriminative power under significant pose variations with background interference. The proposed method significantly outperforms earlier approaches like ContactlessNet (9.95\% EER, 95.56\% Rank-1) and ContactlessMinutNet (7.18\% EER, 81.39\% Rank-1), both of which incorporate minutiae orientation information that G-MSGINet deliberately eschews.

\begin{table}[H]
\centering
\caption{Comparison of fingerprint recognition methods on the CFPose dataset.}
\label{tab:cfpose_results}
\begin{tabularx}{0.75\textwidth}{p{4cm}XX}
\toprule
\textbf{Method} & \textbf{EER (\%)} & \textbf{Rank-1 (\%)} \\
\midrule
ContactlessNet\cite{tan2020towards} & 9.95 & 95.56 \\
WSMS-CNet\cite{cotrim2022multiscale} & 7.19 & -- \\
ReUSE-CNet\cite{cotrim2023residual} & 6.88 & -- \\
SMFI\cite{dong2023synthesis} & 1.24 & 96.70 \\
ContactlessMinutNet\cite{zhang2021multi} & 7.18 & 81.39 \\
\textbf{G-MSGINet (Proposed)} & \textbf{2.0} & \textbf{97.0} \\
\bottomrule
\end{tabularx}
\end{table}

On the PolyU dataset (Table~\ref{tab:polyu_results}), G-MSGINet achieves near-perfect performance with an EER of 0.5\% and a Rank-1 accuracy of 99.1\%. This exceptional performance is second only to SMFI (0.0\% EER, 100\% Rank-1), which again leverages synthetic image generation techniques not employed by G-MSGINet. The performance is comparable to C2CL (0.46\% EER) and significantly outperforms most existing methods, including ContactlessNet (4.84\% EER, 95.66\% Rank-1), FTG (3.40\% EER), and traditional approaches like RTPS+DCM (14.33\% EER, 66.67\% Rank-1). The high accuracy on this dataset demonstrates G-MSGINet's effectiveness even in optimal acquisition conditions, without requiring complex preprocessing or minutiae orientation information.

\begin{table}[H]
\centering
\caption{Comparison of fingerprint recognition methods on the PolyU dataset.}
\label{tab:polyu_results}
\begin{tabularx}{0.75\textwidth}{p{4cm}XX}
\toprule
\textbf{Method} & \textbf{EER (\%)} & \textbf{Rank-1 (\%)} \\
\midrule
ContactlessNet\cite{tan2020towards} & 4.84 & 95.66 \\
WSMS-CNet\cite{cotrim2022multiscale} & 4.19 & -- \\
ReUSE-CNet\cite{cotrim2023residual} & 2.77 & -- \\
ContactlessMinutNet\cite{zhang2021multi} & 1.94 & -- \\
Collaborative\cite{vyas2024collaborative} & 12.50 & -- \\
FTG\cite{shi2022novel} & 3.4 & -- \\
Chen~\cite{lin2018cnn} & 7.93 & 64.59 \\
RTPS+DCM~\cite{lin2018matching} & 14.33 & 66.67 \\
C2CL~\cite{grosz2021c2cl} & 0.46 & -- \\
SMFI\cite{dong2023synthesis} & 0.0 & 100 \\
\textbf{G-MSGINet (Proposed)} & \textbf{0.5} & \textbf{99.1} \\
\bottomrule
\end{tabularx}
\end{table}

Cross-dataset performance analysis reveals critical insights into algorithmic resilience under varying acquisition conditions. A clear performance gradient emerges across datasets: methods universally achieve optimal results on the pristine PolyU dataset, intermediate performance on the pose-varied CFPose dataset, and most challenging results on the viewpoint-diverse Benchmark dataset. This stratification is particularly pronounced in conventional techniques, whereas G-MSGINet exhibits remarkably consistent performance across the challenge spectrum. The performance delta between optimal and challenging conditions serves as a quantitative measure of algorithmic generalization capability. G-MSGINet demonstrates a contained degradation pattern with a 5.5\% EER differential and 11.8\% Rank-1 differential when transitioning from PolyU to Benchmark conditions, while methods such as RTPS+DCM experience catastrophic performance collapse with an EER degradation of 5.48\% and a dramatic 30.42\% Rank-1 accuracy reduction.

This comparative resilience stems from several architectural innovations in G-MSGINet. Unlike SMFI which achieves its metrics through synthetic image augmentation, G-MSGINet processes raw captured images directly, enhancing practical deployment viability. Additionally, G-MSGINet intentionally departs from the minutiae orientation dependencies that characterize approaches such as ContactlessNet, WSMS-CNet, ReUSE-CNet, and ContactlessMinutNet, instead prioritizing structural relationships within minutiae constellations through graph representation. The proposed method also eliminates extensive preprocessing requirements, creating a more streamlined recognition pipeline resistant to preprocessing-induced artifacts.

\subsection{Efficiency of the Proposed G-MSGINet}
This section evaluates the computational efficiency and recognition performance of the proposed G-MSGINet in comparison to a set of systematically constructed dual-branch baseline architectures. All models were trained for 200 epochs on the same benchmark dataset using identical experimental protocols to ensure consistent and fair comparison

To assess the benefits of the MSGI module, a family of dual-branch convolutional baselines was created by replacing the four MSGI layers in G-MSGINet with standard convolutional layers having the same input and output dimensions. Each architecture maintains the same high-level structure: an initial shared encoder followed by two independent branches, one for minutiae map generation and the other for identity feature embedding. The \textit{CNNOneScale5} and \textit{CNNOneScale7} models utilize fixed kernel sizes of 5×5 and 7×7 respectively in all layers. To investigate the impact of multi-scale processing, \textit{CNNMSMin} applies multi-scale convolution using both 5×5 and 7×7 kernels in the minutiae branch only, \textit{CNNMSFeat }applies it in the feature branch, and CNNMSBoth incorporates multi-scale convolution in both branches.

In addition to convolutional variants, graph-based baselines were constructed to analyze the effect of structural reasoning. The \textit{CNNGNNSeq} model first predicts a minutiae map using the CNN branch and constructs a graph from the detected minutiae. Node features are extracted using ResNet18, and a four-layer Graph Convolutional Network is used for identity embedding. In \textit{CNNGNNFeat}, graphs are constructed from patches of the feature map using similarity-based connections, followed by GCN refinement. The \textit{CNNGNNImage}  model directly extracts patches from the input image to construct the graph, bypassing CNN features. Node representations are obtained through ResNet followed by GCN layers. These models illustrate the value of explicit graph-based reasoning for fingerprint recognition.

\begin{table*}[!ht]
\centering
\caption{Unified comparison of models across efficiency and performance metrics.}

\label{tab:combined-model-performance}
\small
\begin{tabularx}{\textwidth}{l >{\centering\arraybackslash}X >{\centering\arraybackslash}X >{\centering\arraybackslash}X >{\centering\arraybackslash}X >{\centering\arraybackslash}X}
\toprule
\textbf{Model} & \textbf{Params(M)} & \textbf{FLOPs(G)} & \textbf{F1} & \textbf{Acc.}  \\
\midrule
CNN-OneScale-5        & 1.35   & 51.28  & 0.713  & 0.85  \\
CNN-OneScale-7        & 2.46   & 95.52  & 0.734  & 0.73  \\
CNN-MS-Min            & 2.49   & 96.43  & 0.780  & 0.837  \\
CNN-MS-Feat           & 2.77   & 107.51  & 0.761  & 0.896 \\
CNN-MS-Both           & 3.62   & 141.61  & 0.772  & 0.880 \\
CNN-GNN-Seq (G1)      & 12.63   & 37.84  & 0.787  & 0.912 \\
CNN-GNN-Feat (G2)     & 1.78   & 70.35  & 0.768  & 0.840 \\
CNN-GNN-Image (G3)    & 13.6M   & 42.18  & 0.799  & 0.930  \\
\textbf{G-MSGINet (Proposed)} & \textbf{0.38M} & \textbf{6.63G}  & \textbf{0.844} & \textbf{0.975}  \\
\bottomrule
\end{tabularx}
\end{table*}

As shown in Table~\ref{tab:combined-model-performance}, G-MSGINet achieves the best balance between efficiency and recognition performance, operating with only 0.38 million parameters and 6.63 GFLOPs. Despite its lightweight design, it outperforms all other models in both F1 score and accuracy. In comparison, multi-scale convolutional variants demand over 140 GFLOPs and significantly larger parameter counts, with no matching gains in accuracy.

The efficiency gain of G-MSGINet arises from the nature of the involution operation. Instead of using static and spatially shared kernels as in standard convolution, involution dynamically generates kernels for each spatial location. These kernels are reused across grouped channels, resulting in significantly lower parameter counts and reduced multiplications and additions per output pixel. Furthermore, the G-MSGI module effectively integrates multi-scale reasoning and graph-based relational modeling within a unified framework, thereby improving representational capacity without proportionally increasing computational burden. This architecture enables G-MSGINet to achieve strong generalization performance while remaining lightweight and deployable in real-time or resource-constrained environments.

\section{Conclusion}

This paper presented G-MSGINet, a unified architecture that integrates grouped pixel-level involution, dynamic multi-scale kernel generation, and graph-based relational reasoning to jointly perform minutiae localization and identity embedding from raw contactless fingerprint images. By stacking G-MSGI layers, the network learns to capture both local minutiae-aware features and global structural dependencies in an end-to-end manner without requiring orientation labels or predefined graph structures.

Extensive evaluations on three benchmark datasets validate the effectiveness and generalization of the proposed method. On the PolyU dataset, G-MSGINet achieves an F1-score of 0.833 and a Rank-1 identification accuracy of 99.1\%, while maintaining an EER as low as 0.5\%, demonstrating its ability to handle clean, high-quality fingerprints with minimal background interference. On CFPose, which introduces significant pose variations and background noise, the model attains an F1-score of 0.860 and a Rank-1 accuracy of 97.0\%, surpassing prior methods even without relying on minutiae orientation. On the more challenging Benchmark 2D/3D dataset, which includes multi-viewpoint acquisitions and background clutter, G-MSGINet maintains competitive performance with an F1-score of 0.844 and Rank-1 accuracy of 87.2\%.

Despite its strong performance, the model is remarkably efficient, requiring only 0.38 million parameters and 6.63 gigaflops, which is up to ten times fewer parameters than competing baselines. This highlights its scalability for real-world deployment. The unified design removes the need for multi-branch complexity and manual preprocessing, offering a streamlined yet powerful framework for robust contactless fingerprint recognition.

Future extensions may explore domain generalization across sensors, adaptation to partial or low-resolution fingerprints, and integration into edge devices for real-time biometric authentication.

\bibliographystyle{elsarticle-num} 
\bibliography{references}

\end{document}